\pdfoutput=1
\documentclass[twocolumn]{article}
\usepackage{framed,multirow}
\usepackage{amssymb}
\usepackage{latexsym}
\usepackage{url}
\usepackage{xcolor}
\definecolor{newcolor}{rgb}{.8,.349,.1}

\usepackage{amsmath}
\usepackage{booktabs}
\usepackage{graphicx}
\usepackage{pgf-pie}
\usepackage{subcaption}
\usepackage[normalem]{ulem}
\usepackage{authblk}
\usepackage[nameinlink,capitalize]{cleveref}
\usepackage{natbib}

\newcommand{\x}{\mathbf{x}}
\newcommand{\y}{\mathbf{y}}

\DeclareMathOperator*{\argmax}{arg\,max~}
\newcommand{\p}{\mathrm{p}}

\title{Modernizing Historical Documents: a User Study}

\author{Miguel Domingo and Francisco Casacuberta}
\affil{Pattern Recognition and Human Language Technology Research Center \protect\\ Universitat Polit{\`e}cnica de Val{\`e}ncia - Camino de Vera s/n, 46022 Valencia, Spain \protect\\ $ $ \protect\\ \emph{midobal@prhlt.upv.es, fcn@prhlt.upv.es}}
\date{}

\begin{document}
\maketitle

\begin{abstract}
    Accessibility to historical documents is mostly limited to scholars. This is due to the language barrier inherent in human language and the linguistic properties of these documents. Given a historical document, modernization aims to generate a new version of it, written in the modern version of the document's language. Its goal is to tackle the language barrier, decreasing the comprehension difficulty and making historical documents accessible to a broader audience. In this work, we proposed a new neural machine translation approach that profits from modern documents to enrich its systems. We tested this approach with both automatic and human evaluation, and conducted a user study. Results showed that modernization is successfully reaching its goal, although it still has room for improvement.
\end{abstract}

\section{Introduction}
Historical documents are an important part of our cultural heritage. However, the nature of human language, which evolves with the passage of time, and the linguistic properties of these documents{\textemdash}due to the lack of a spelling convention, orthography changes depending on the time period and author{\textemdash}increase the difficulty of comprehending them. For this reason, historical documents are mostly accessible to scholars. Thus, in order to preserve them and make them reachable to a broader audience, a scholar is typically in charge of producing a comprehensive contents document which allows non-experts to locate and gain a basic understanding of a given document \citep[e.g.,][]{Monk18}.

Modernization aims to tackle this language barrier by generating a new version of a historical document, written in the modern version of the document's original language. \cref{fi:Shakespeare} shows an example of modernizing a document. In this case, part of the language structures and rhymes have been lost. However, the modern version is easier to read and comprehend by a broader audience. This problem is also present in poetry translation since the entwinement between sound and word and sense cannot be truly replicated in a different language \citep{Ilonka18}. However, translating a poem from one language into another is a way of sharing cultural practices and ideologies across languages \citep{Rajvanshi15}.

\begin{figure*}[!ht]
	\centering
	\scriptsize
	\begin{minipage}{0.001\textwidth}
		$ $
	\end{minipage}
	\begin{minipage}{0.42\textwidth}
		O Romeo, Romeo! Wherefore art thou Romeo?
		
		Deny thy father and refuse thy name.
		
		Or, if thou wilt not, be but sworn my love,
		
		And I'll no longer be a Capulet.
		
		$ $
		
		With love's light wings did I o'erperch these walls,
		
		For stony limits cannot hold love out,
		
		And what love can do, that dares love attempt.
		
		Therefore thy kinsmen are no stop to me.
	\end{minipage}
	\begin{minipage}{0.001\textwidth}
		$ $
	\end{minipage}
	\begin{minipage}{0.56\textwidth}
		Oh, Romeo, Romeo, why do you have to be Romeo? 
		
		Forget about your father and change your name. 
		
		Or else, if you won't change your name, just swear you love me 
		
		and I'll stop being a Capulet.
		
		$ $
		
		I flew over these walls with the light wings of love. 
		
		Stone walls can't keep love out. 
		
		Whatever a man in love can possibly do, his love will make him try to do it. 
		
		Therefore your relatives are no obstacle.
	\end{minipage}
	\caption{Example of modernizing a historical document. The original text is composed of fragments from \emph{Romeo and Juliet} by \emph{William Shakespeare}. The modernized version was obtained from \cite{Crowther03b}.}
	\label{fi:Shakespeare}
\end{figure*}

Modernization can be a controversial topic since it implies an alteration of the original document (e.g., the manual modernization of \emph{El Quijote} rose a controversy in Spain \citep{Flood15}). However, it is manually applied to classic literature in order to make works that had been relegated to scholars due to the hardness of their comprehension, understandable to contemporary readers \citep{Rodriguez15}.

Finally, while the language richness present in historical documents is also part of our cultural heritage, the goal of modernization is to make historical documents accessible to a general audience. Other research topics on historical document are focused on different aspects of their richness. For example, historical manuscripts are automatically transcribed and digitized \citep{Toselli10,Toselli17}. Orthography is normalized to account for the lack of a spelling convention \citep{Laing93, Porta13}. Search queries can find all occurrences of one or more words \citep{Rogers91,Ernst06}. Word frequency lists are generated \citep[e.g.,][]{Baron09}. And natural language processing tools provide automatic annotations to identify and extract linguistic structures such as relative clauses \citep{Hundt11} or verb phrases \citep{Fiebranz11, Pettersson13}.

In this work, we followed a machine translation (MT) approach to tackle the modernization problem. Similarly to \citet{Domingo18b}, we profited from modern documents to enrich the modernization systems. However, we applied a data selection technique to take better profit of these documents, selecting only the most relevant sentences for each task. We evaluated our approach both automatically and with the help of 4 scholars specialized in classic Spanish literature. Additionally, we conducted a user study with 42 people to assess whether or not modernization is able to decrease the difficulty of comprehending historical documents. Our main contributions are as follows:

\begin{itemize}
	\item We proposed a new neural machine translation (MT) approach that successfully profits from modern documents to enrich its modernization systems.
	\item We tested our proposal using 3 datasets from different languages and time periods.
	\item We assessed the quality of our proposal using both automatic and human evaluation, conducted by 4 scholars specialized in classic Spanish literature.
	\item First time, to the best of our knowledge, in which an NMT modernization approach behaves similarly or better than a statistical machine translation (SMT) modernization approach.
	\item We conducted a study with 42 users to assess whether modernization successfully decreases the difficulty of comprehending historical documents.
\end{itemize}

The rest of this document is structured as follows: \cref{se:work} introduces the related work. Then, \cref{se:mod} presents the modernization approach. After that, in \cref{se:exp}, we describe the experimental framework of our work. Next, in \cref{se:eval}, we present and discuss the evaluation conducted in order to assess our approach. \cref{se:study} describes and presents the user study. Finally, in \cref{se:con}, conclusions are drawn.

\section{Related work}
\label{se:work}

Modernization has been manually applied to literature for centuries. One of the most well-known examples is \emph{The Bible}, which has been adapted and translated for generations in order to preserve and transmit its contents \citep{Given15}. Classic literature is also frequently modernized in order to bring it closer to a contemporary audience (e.g., \emph{No Fear Shakespeare}\footnote{\url{https://www.sparknotes.com/shakespeare/}.}; \emph{Odres Nuevos}\footnote{\url{https://www.castalia.es/libros?tipo=coleccion&letra=O&nombre=49&other_page=1}.}; \emph{El Quijote} \citep{Trapiello15}). However, on the literature we find that, while normalizing orthography to account for the lack of a spelling convention has been extensively research for years \citep{Laing93,Baron08,Porta13,Hamalainen18}, automatic modernization of historical documents is a young research field. 

One of the first related works was a shared task for translating historical text to contemporary language \citep{Sang17}. The task was focused on normalizing the document's spelling. However, they also approached document modernization using a set of rules.  \citet{Domingo17a} proposed a modernization approach based on SMT. An  NMT approach was proposed by \citet{Domingo18b}. Finally, \citet{Sen19} augmented the training data by extracting pairs of phrases and adding them as new training sentences.

\section{Modernization approaches}
\label{se:mod}
In this section, we present the state-of-the-art SMT modernization approach and our NMT-based proposal. Both approaches rely on MT which, given a source sentence $\x$, aims at finding the most likely translation $\hat{\y}$ \citep{Brown93}:

\begin{equation}
\hat{\y} = \argmax_{\y} Pr(\y \mid \x)
\label{eq:smt}
\end{equation}

\subsection{SMT approach}
\label{se:smt}
For years, SMT has been the prevailing approach to compute \cref{eq:smt}, using models that rely on a log-linear combination of different models \citep{Och02}: namely, phrase-based alignment models, reordering models and language models; among others \citep{Zens02,Koehn03}.

In this approach, modernization is tackled as a conventional translation task: training an SMT system from a parallel corpora in which, for each sentence of the original document, its corresponding modernized version is available. For training this system, the language of the original document is considered as the source language, and its modernized version as the target language.

\subsection{NMT approach}
\label{se:nmt}
NMT models \cref{eq:smt} with a neural network which usually follows an encoder-decoder architecture, in which the source sentence is projected into a distributed representation at the encoding step. Then, at the decoding step, the decoder generates its most likely translation{\textemdash}word by word{\textemdash}using a beam search method \citep{Sutskever14}. 

The system's input is a word sequence in the source language. An embedding matrix linearly projects each word to a fixed-size real-valued vector. These words embeddings are, then, fed into a bidirectional \citep{Schuster97} long short-term memory (LSTM) \citep{Hochreiter97} network. As a result, a sequence of annotations is produced by concatenating the hidden states from the forward and backward layers. An attention mechanism \citep{Bahdanau15} allows the decoder to focus on parts of the input sequence, computing a weighted mean of annotated sequences. A soft alignment model computes these weights, weighting each annotation with the previous decoding state. Another LSTM network{\textemdash}conditioned by the representation computed by the attention model and the last word generated{\textemdash}is used for the decoder. Finally, a distribution over the target language vocabulary is computed by the deep output layer \citep{Pascanu13}. The model is trained by applying stochastic gradient descent jointly to maximize the log-likelihood over a bilingual parallel corpus.

As the SMT approach (see \cref{se:smt}), our proposal tackles modernization as a conventional translation task but using NMT instead of SMT. Additionally, since NMT systems need larger quantities of training data, and a frequent problem when working with historical documents is the scarce availability of parallel training data \citep{Bollmann16}, we created synthetic data in order to profit from modern documents to enrich the NMT models. First, we applied feature decay algorithm \citep{Biccici15} to select those documents which are closer to the ones we have to modernize. After that, we followed a backtranslation approach \citep{Sennrich15} to create a parallel synthetic corpus. Backtranslation has become the norm when building state-of-the-art  NMT  systems{\textemdash}especially in resource-poor scenarios \citep{Poncelas18}. Given a monolingual corpus in the target language and an MT system trained to translate from the target language to the source language, the synthetic data is generated by translating the monolingual corpus with the MT system{\textemdash}the resulting data is used as the source part of the corpus, and the monolingual data as the target part.

\section{Experimental framework}
\label{se:exp}
\begin{table*}[!ht]
	\caption{Corpora statistics. $|S|$ stands for number of sentences, $|T|$ for number of tokens and $|V|$ for size of the vocabulary. \emph{Modern documents} refers to the monolingual data used to create the synthetic data. M denotes millions and K thousands.}
	\label{ta:corp}
	\centering
	\resizebox{0.75\textwidth}{!}{\begin{minipage}{\textwidth}
			\centering
			\begin{tabular}{c c c c c c c c}
				\toprule
				&  & \multicolumn{2}{c}{\textbf{Dutch Bible}} & \multicolumn{2}{c}{\textbf{El Quijote}} & \multicolumn{2}{c}{\textbf{OE-ME}} \\
				\cmidrule(lr){3-4}\cmidrule(lr){5-6}\cmidrule(lr){7-8}
				&  & \textbf{Original} & \textbf{Modernized} & \textbf{Original} & \textbf{Modernized} & \textbf{Original} & \textbf{Modernized} \\
				\midrule
				\multirow{3}{*}{Train} & $|S|$ & \multicolumn{2}{c}{35.2K} & \multicolumn{2}{c}{10K} & \multicolumn{2}{c}{2716} \\
				& $|T|$ & 870.4K & 862.4K & 283.3K & 283.2K & 64.3K & 69.6K \\
				& $|V|$ & 53.8K & 42.8K & 31.7K & 31.3K & 13.3K & 8.6K \\
				\midrule
				\multirow{3}{*}{Validation} & $|S|$ & \multicolumn{2}{c}{2000} & \multicolumn{2}{c}{2000} & \multicolumn{2}{c}{500} \\
				& $|T|$ & 56.4K & 54.8K & 53.2K & 53.2K & 12.2K & 13.3K \\
				& $|V|$ & 9.1K & 7.8K & 10.7K & 10.6K & 4.2K & 3.2K \\
				\midrule
				\multirow{3}{*}{Test} & $|S|$ & \multicolumn{2}{c}{5000} & \multicolumn{2}{c}{2000} & \multicolumn{2}{c}{500} \\
				& $|T|$ & 145.8K & 140.8K & 41.8K & 42.0K & 11.9K & 12.9K \\
				& $|V|$ & 10.5K & 9.0K & 8.9K & 9.0K & 4.1K & 3.2K \\
				\midrule
				\multirow{3}{*}{Modern documents} & $|S|$ & \multicolumn{2}{c}{3.0M} & \multicolumn{2}{c}{2.0M} & \multicolumn{2}{c}{6.0M} \\
				& $|T|$ & 76.1M & 74.1M & 22.3M & 22.2M & 67.5M & 71.6M \\
				& $|V|$ & 1.7M & 1.7M & 210.1K & 211.7K & 290.2K & 287.4K \\
				\bottomrule
			\end{tabular}
	\end{minipage}}
\end{table*}

In this section, we describe the MT systems, corpora and evaluation metrics from our experimental framework.

\subsection{MT systems}
SMT systems were trained with \texttt{Moses} \citep{Koehn07}, following the standard procedure: we estimated a 5-gram language model{\textemdash}smoothed with the improved KneserNey method{\textemdash}using \texttt{SRILM} \citep{Stolcke02}, and optimized the weights of the log-linear model with MERT \citep{Och03a}. SMT systems were used both for the SMT modernization approach and for generating synthetic data (see \cref{se:mod}).

We built NMT systems using \texttt{OpenNMT-py} \citep{Klein17}. We used long short-term memory units \citep{Gers00}, with all model dimensions set to $512$. We trained the system using Adam \citep{Kingma14} with a fixed learning rate of $0.0002$ and a batch size of $60$. We applied label smoothing of $0.1$ \citep{Szegedy15}. At inference time, we used beam search with a beam size of 6. In order to reduce vocabulary, we applied joint byte pair encoding (BPE) \citep{Sennrich16} to all corpora, using $32,000$ merge operations. NMT systems were trained using synthetic data and, then, were fine-tuned with the training data.

\subsection{Corpora}
\begin{description}
	\item[Dutch Bible] \citep{Sang17}: A collection of different versions of the Dutch Bible. Among others, it contains a version from 1637{\textemdash}which we consider as the original version{\textemdash}and another from 1888{\textemdash}which we consider as the modern version (using 19$^{\mathrm{th}}$ century Dutch as if it were \emph{modern Dutch}).
	
	\item[El Quijote] \citep{Domingo18b}: the well-known 17$^{\mathrm{th}}$ century Spanish novel by Miguel de Cervantes, and its correspondent 21$^{\mathrm{st}}$ century version.
	
	\item[OE-ME] \citep{Sen19}: contains the original 11$^{\mathrm{th}}$ century English text \emph{The Homilies of the Anglo-Saxon Church} and a 19$^{\mathrm{th}}$ century version{\textemdash}which we consider as \emph{modern English}.
\end{description}

As reflected in \cref{ta:corp}, the corpora sizes are small. Thus, the use of synthetic data to profit from modern documents and increase the training data (see \cref{se:nmt}). As \emph{modern documents}, we made use of the collection of Dutch books available at the \emph{Digitale Bibliotheek voor de Nederlandse letteren}\footnote{\url{http://dbnl.nl/}.}, for Dutch; and OpenSubtitles \citep{Lison16}{\textemdash}a collection of movie subtitles in different languages{\textemdash}for Spanish and English. 

\subsection{Metrics}
Modernization adopted evaluation metrics from MT. In order to assess our proposal, we made use of:

\begin{description}
	\item[Translation Error Rate (TER)] \citep{Snover06}: number of word edit operations (insertion, substitution, deletion and swapping), normalized by the number of words in the final translation.
	\item[BiLingual Evaluation Understudy (BLEU)]~\citep{Papineni02}: geometric average of the modified n-gram precision, multiplied by a brevity factor.
\end{description}

We used \texttt{sacreBLEU} \citep{Post18} in order to ensure consistent BLEU scores. Additionally, we applied approximate randomization tests \citep{Riezler05}{\textemdash}with $10,000$ repetitions and using a $p$-value of $0.05${\textemdash}to determine whether two systems presented statistically significance.

\section{Evaluation}
\label{se:eval}
In order to assess the quality of our modernization approaches, we started by performing an automatic evaluation. Then, with the help of 4 scholars, we conducted a human evaluation.

\subsection{Automatic evaluation}
\label{se:aev}
\begin{table*}[!ht]
	\caption{Experimental results. \emph{Baseline} system corresponds to considering the original document as the modernized version. $^\dagger$ indicates statistically significance between the SMT/NMT approaches and the baseline. $^\ddagger$ indicates statistically significance between the NMT and SMT approaches. [$\downarrow$] indicates that the lowest the value the highest the quality. [$\uparrow$] indicates that the highest the value the highest the quality.}
	\label{ta:res}
	\centering
	\resizebox{0.75\textwidth}{!}{\begin{minipage}{\textwidth}
			\centering
			\begin{tabular}{l c c c c c c }
				\toprule
				\multirow{2}{*}{\textbf{Approach}} & \multicolumn{2}{c}{\textbf{Dutch Bible}} & \multicolumn{2}{c}{\textbf{El Quijote}} & \multicolumn{2}{c}{\textbf{OE-ME}} \\
				\cmidrule(lr){2-3}\cmidrule(lr){4-5}\cmidrule(lr){6-7}
				& TER [$\downarrow$] & BLEU [$\uparrow$] & TER [$\downarrow$] & BLEU [$\uparrow$] & TER [$\downarrow$] & BLEU [$\uparrow$] \\
				\midrule
				Baseline & $57.9$ & $12.9$ & $44.2$ & $36.3$ & $91.0$ & $2.8$ \\
				\midrule
				SMT & $11.5^\dagger$ & $77.5^\dagger$ & $30.7^\dagger$ & $58.3^\dagger$ & $39.6^\dagger$ & $39.6^\dagger$ \\
				NMT & $11.1^{\dagger\ddagger}$ & $80.6^{\dagger\ddagger}$ & $31.9^\dagger$ & $57.3^\dagger$ & $44.3^\dagger$ & $35.9^\dagger$ \\
				\bottomrule
			\end{tabular}
	\end{minipage}}
\end{table*}

\cref{ta:res} presents the results of the experimental session. All approaches significantly improved the modernization quality. Differences between the SMT and NMT approaches were only statistically significant for \emph{Dutch Bible}. In that case, the NMT approach yielded the best results: an overall improvement of $46.8$ points according to TER and $67.7$ points according to BLEU; and an improvement of $0.4$ and $2.9$ points according to TER and BLEU respectively, with respect to the SMT approach.

To the best of our knowledge, this is the first time that an NMT modernization approach is able to achieve these kinds of results. \citet{Domingo18b} already tried to profit from modern documents to enrich the neural models. However, their approach only improved the modernization quality in some cases{\textemdash}and never enough to reach the quality of the SMT approach{\textemdash}while in others it lowered it significantly. Our approach was based on theirs, but we used a data selection technique to help us filtered the monolingual data in order to generate synthetic data more suitable for each task.

\subsection{Human evaluation}
\label{se:heval}
The human evaluation was performed by 4 scholars specialized in classic Spanish literature. For this reason, it was conducted using \emph{El Quijote}. We randomly selected 100 sentences, checking that modernizations were different to the original sentences. We showed each sentence together with its modernization{\textemdash}50 sentences modernized with the SMT approach and another 50 with the NMT approach{\textemdash} and asked the scholars to give a rating according to the quality of the following aspects: fluency, lexical meaning, syntax, semantic and modernization. To avoid any bias, we shuffled the sentences and did not give any detail to the evaluators about how modernizations had been produced. \cref{ta:scholars} shows the results of the evaluation.

\begin{table*}[!ht]
	\caption{Results of the human evaluation. Values correspond to the average score for all sentences of each approach. 1 is the lowest score and 5 is the highest.}
	\label{ta:scholars}
	\resizebox{0.65\textwidth}{!}{\begin{minipage}{\textwidth}
			\begin{tabular}{l c c c c c c c c c c}
				\toprule
				\multirow{2}{*}{\textbf{Scholar}} & \multicolumn{5}{c}{\textbf{SMT approach}} & \multicolumn{5}{c}{\textbf{NMT approach}} \\
				\cmidrule(lr){2-6}\cmidrule(lr){7-11}
				& \textbf{Fluency} & \textbf{Lexical meaning} & \textbf{Syntax} & \textbf{Semantic} & \textbf{Modernization} & \textbf{Fluency} & \textbf{Lexical meaning} & \textbf{Syntax} & \textbf{Semantic} & \textbf{Modernization} \\
				\midrule
				Scholar$_\mathrm{1}$ & $5.0$ & $4.3$ & $4.3$ & $4.6$ & $3.9$ & $4.8$ & $4.0$ & $4.0$ & $4.1$ & $4.0$ \\
				Scholar$_\mathrm{2}$ & $2.1$ & $1.9$ & $2.0$ & $2.1$ & $2.0$ & $2.0$ & $1.9$ & $1.9$ & $1.9$ & $1.9$ \\
				Scholar$_\mathrm{3}$ & $3.2$ & $3.1$ & $2.9$ & $2.9$ & $3.1$ & $3.3$ & $3.2$ & $2.9$ & $3.0$ & $3.1$ \\
				Scholar$_\mathrm{4}$ & $4.5$ & $3.9$ & $4.6$ & $4.3$ & $4.0$ & $3.8$ & $3.5$ & $3.7$ & $3.7$ & $3.5$ \\
				\midrule
				Average & $3.7$ & $3.3$ & $3.4$ & $3.5$ & $3.2$ & $3.4$ & $3.1$ & $3.1$ & $3.2$ & $3.1$ \\
				\bottomrule
			\end{tabular}
	\end{minipage}}
\end{table*}

While the automatic evaluation (see \cref{se:aev}) did not show any significant differences between the SMT and NMT approaches, the human evaluators slightly preferred SMT over NMT. Scores vary considerably depending on the evaluator{\textemdash}scholar$_\mathrm{1}$ and scholar$_\mathrm{4}$ gave higher scores than scholar$_\mathrm{2}$ and scholar$_\mathrm{3}$. However, all evaluators agreed that fluency is the strongest point of both approaches. In general, scores are above the average, which seems to correlate with the automatic evaluation.

When we asked evaluators about their opinion, they commented that the main problems were related with punctuation and diacritical marks. They also mentioned that, sometimes, part of the sentence was lost in the modernization{\textemdash}a known issue related with NMT \citep{Wu16}. Additionally, scholar$_\mathrm{1}$ commented that, overall, the quality of the modernization was acceptable. However, scholar$_\mathrm{2}$ commented that if they had to correct the mistakes, they would have preferred to do the modernization from scratch.

\section{User study}
\label{se:study}
\begin{figure*}[!ht]
	\begin{minipage}{0.4\textwidth}
		\begin{tikzpicture}
		\tikzstyle{every node}=[font=\tiny]
		\pie[radius=1.7,color={black!10,black!20,black!30,black!40,black!50,black!60}]{2.4/$\le$ 20 years, 33.3/21{\textendash}30 years, 26.2/31{\textendash}40 years, 21.4/41{\textendash}50 years, 9.5/51{\textendash}60 years, 7.1/61{\textendash}70 years}
		\end{tikzpicture}
		\subcaption{Age distribution.}\label{fi:age}
	\end{minipage}
	\begin{minipage}{0.5\textwidth}
		\begin{tikzpicture}
		\tikzstyle{every node}=[font=\tiny]
		\pie[radius=1.7,color={black!10,black!20,black!30,black!40,black!50,black!60,black!70},rotate=0]{2.4/Unfamiliar, 7.1/Know what it is about, 14.3/Read fragments of an adaptation, 19.0/Read an adaptation, 14.3/Read fragments of the original, 35.7/Read the original, 7.1/Read a modernized version}
		\end{tikzpicture}
		\subcaption{Familiarity with \emph{El Quijote}.}\label{fi:quijote}
	\end{minipage}
	\caption{Information about study participants.}
	\label{fi:users}
\end{figure*}
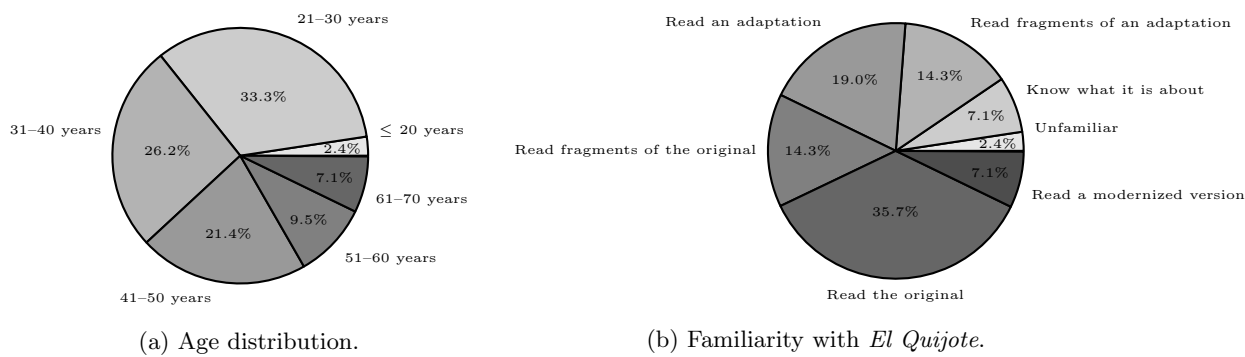

In order to assess whether modernization is able to decrease the difficulty of comprehending historical documents and, thus, making them accessible to a broader audience, we conducted a user study using \emph{El Quijote}. 42 participants took part in this study. Considering that \emph{El Quijote} is well-known in Spain, we asked participants about their familiarity with it. \cref{fi:users} shows some information about the user's age and their familiarity with \emph{El Quijote}.

The majority of the participants were between 20 and 50 years old, but there were also older and younger people. With one exception, all participants were familiar with \emph{El Quijote} to some extent. In fact, 35.7\% of them had read the original version of the novel.

The study consisted in several questions in which we showed two sentences to the user{\textemdash}the original sentence and its modernized version (either by the SMT or the NMT approach){\textemdash}and asked them to select which sentence was easier for them to read and comprehend, if both of them had the same difficulty, or if they thought that both sentence did not have the same meaning. The selected sentences were the same used in the human evaluation (see \cref{se:heval}). In order to avoid any bias, the order in which sentences appeared (i.e., the original sentence and its modernized version) was randomized, as well as the use of the different approaches. \cref{fi:exa} shows an example of a question.

\begin{figure*}[!ht]
	\small
	\textbf{Select the sentence which is easier for you to read and comprehend:}
	\begin{itemize}
		\item Y, leuantandose, dex{\'o} de comer, y fue a quitar la cubierta de la primera imagen, que mostro ser la de San Iorge puesto a cauallo, con vna serpiente enroscada a los pies, y la lan\c{c}a atrauessada por la boca, con la fiere\c{c}a que suele pintarse.
		\item Y levant{\'a}ndose, dej{\'o} de comer, y fue a quitar la cubierta de la primera imagen, que mostr{\'o} ser la de San Jorge puesto a caballo, con una serpiente enroscada a los pies, y la lanza atravesada por la boca, con la fiere\c{c}a que suele pintarse.
		\item Indifferent.
		\item Both sentences do not have the same meaning.
	\end{itemize}
	\caption{Example of a question.}
	\label{fi:exa}
\end{figure*}

\cref{ta:study} presents the results of the study. Despite the users' familiarity with \emph{El Quijote}, modernization succeed in making it easier to comprehend. No matter the modernization approach, users selected the modernized version in the majority of the cases. In most of the remaining cases, users did not find any significant difference with respect to the original sentence.

When comparing both approaches, we observe that the SMT approach yielded better results: Users selected 61.4\% of their modernized versions, while they only selected a 50.9\% of the sentences modernized by the NMT approach. Additionally, the SMT approach only introduced errors in 7.8\% of the cases{\textemdash}the NMT introduced them in 20.3\% of the cases{\textemdash}and its modernized versions were harder to comprehend only in 3.2\% of the cases{\textemdash}versus a 6.4\% of the cases for the NMT approach. Therefore, despite neither the automatic nor the human evaluation were able to find significant differences between both approaches, the user study showed that the SMT approach produced versions easier to read and comprehend more successfully than the NMT approach.

\begin{table*}[!ht]
	\caption{Results [\%] of the user study. \emph{Original} means that users understood better the original version. \emph{Modernized} means that users understood better the modernized version. \emph{Indifferent} means that users did not found any significant differences between the original and modernized versions. \emph{Not equal} means that users felt that the meaning between both version differ.}
	\label{ta:study}
	\centering
	\resizebox{0.75\textwidth}{!}{\begin{minipage}{\textwidth}
			\centering
			\begin{tabular}{c c c c c c c c}
				\toprule
				\multicolumn{4}{c}{\textbf{SMT}} & \multicolumn{4}{c}{\textbf{NMT}} \\
				\cmidrule(lr){1-4}\cmidrule(lr){5-8}
				\textbf{Original} & \textbf{Modernized} & \textbf{Indifferent} & \textbf{Not equal} & \textbf{Original} & \textbf{Modernized} & \textbf{Indifferent} & \textbf{Not equal} \\
				\midrule
				$3.2$ & $61.4$ & $27.6$ & $7.8$ & $6.4$ & $50.9$ & $22.3$ & $20.3$ \\
				\bottomrule
			\end{tabular}
	\end{minipage}}
\end{table*}

\subsection{Qualitative analysis}
In this section, we show some behavioral examples of the modernization approach. The example from \cref{fi:exa} shows a successfully modernized sentence. Except for one small mistake (\emph{fiere\c{c}a}, which should be \emph{fiere\textbf{z}a}), orthography has been successfully modernized, making the sentence easier to read. (Note that, in this case, orthography is the only thing that needs to be modified in order to achieve a modern Spanish version.)

\begin{figure*}[!ht]
	\small
	\textbf{Original version:} Huuolo de conceder don Quixote, y assi lo hizo. \\
	\textbf{Modernized version:} Hu{\'e}olo de conceder don Quijote, y as{\'i} lo hizo.
	\caption{Example of modernization in which the modernized version is similar to the original version.}
	\label{fi:exiq}
\end{figure*}

\cref{fi:exiq} shows an example in which there is not any significant difference between the modernized and the original version. Only three words have been modified{\textemdash}and one of them (\emph{hu{\'e}olo}) is not even a real word but a mistake introduced by the use of BPE. Despite this, there are people who found the modernized version easier to read; a great majority that found no difference between them; and a few people that either preferred the original version or considered that they did not have the same meaning.

\begin{figure*}[!ht]
	\small
	\textbf{Original version:} Ofreciosele el gallardo pastor, pidiole que se viniesse con el a sus tiendas; \\
	\textbf{Modernized version:} Se le ros{\'o} el gallardo pastor, pile dio que se viniese con {\'e}l a sus tiendas;
	\caption{Example of modernization in which users preferred the original version over the modernized one.}
	\label{fi:exorg}
\end{figure*}

In \cref{fi:exorg}, we can see an example in which the original sentence is easier to understand than its modernized version. While users considered both versions to have the same meaning, the modernized one is harder to comprehend since the first half of the sentence does not make much sense. In fact, looking at the human evaluation, scholars considered the modernized version to be more or less fluent, but with a poor lexical meaning, syntax and semantic.

\begin{figure*}[!ht]
	\small
	\textbf{Original version:} Que me maten si los encantadores que me persiguen no quieren enredarme en ellas, y detener mi camino, como en vengança de la riguridad que con Altissidora he tenido. \\
	\textbf{Modernized version:} – Con mucho gusto?
	\caption{Example of modernization in which the modern version differs with respect to the original.}
	\label{fi:exdif}
\end{figure*}

Finally, \cref{fi:exdif} shows an example in which the modernization went very bad. On the one hand, the modernized version is way shorter than the original version. On the other hand, its meaning has no relation with the original one.

\section{Conclusions and future work}
\label{se:con}
In this work, we proposed a new NMT modernization approach in order to tackle the language barrier inherent in historical documents. We tested this approach on three different historical datasets from three different languages and time periods, comparing it with the state-of-the-art SMT approach.

An automatic evaluation showed that our approach improved the results achieved by the SMT approach on one dataset. Results were not statistically different than the SMT ones for the other two datasets. Additionally, we conducted a human evaluation for the Spanish dataset. This evaluation involved 4 scholars specialized in classical Spanish literature. Its results correlated with the automatic evaluation.

Finally, we conducted a user study to evaluate whether modernization{\textemdash}both SMT and NMT approaches{\textemdash}was able to decrease the difficulty of comprehending historical documents and, thus, increase their accessibility to a broader audience. 42 volunteers, of different age and background, participated in this study. The study was conducted using the same Spanish subset than for the human evaluation. Results showed that modernization successfully decreased the comprehension difficulty. In most of the cases, users chose the modernized version as the easiest to read and comprehend. However, there is still room for improvement. Sometimes, the modernization introduced errors that made users feel that the meaning had been changed. Other times, users did not find any significant difference between the original version and its modernization. When comparing the SMT and NMT approaches, the NMT approach made a bigger number of errors and the user chose its modernized versions as the best option fewer times than with the SMT approach.

While results showed that modernization had successfully improved the understanding of historical documents, we have to take into consideration that language-related losses may appear during the process (e.g., \cref{fi:Shakespeare} shows an example in which part of the language structures and rhymes disappear). However, the goal of modernization is limited to bringing understanding of historical documents to a general audience.

As a future work, we would like to tackle the main problems pointed out during the human evaluation and the user study. Mainly, punctuation, diacritical marks, the introduction of non-existent words and loosing part of the sentence. We would also like to conduct a new human evaluation involving more scholars and more languages and datasets, and a new user study for different languages and datasets. Finally, we would like to apply the field of interactive machine translation to modernization, in order to assist scholars to achieve an error-free modernization.

\section*{Acknowledgments}
The authors wish to thank the anonymous reviewers for their careful reading and in-depth criticisms and suggestions. The research leading to these results has received funding from the European Union through \emph{Programa Operativo del Fondo Europeo de Desarrollo Regional (FEDER)} from Comunitat Valenciana (2014{\textendash}2020) under project \emph{Sistemas de frabricaci{\'o}n inteligentes para la ind{\'u}stria 4.0} (grant agreement IDIFEDER/2018/025); from  Ministerio de Econom{\'i}a y Competitividad (MINECO) under project \emph{MISMIS-FAKEnHATE} (grant agreement PGC2018-096212-B-C31); from Fundaci{\'o}n BBVA under project \emph{Carabela} (grant agreement CARABELA); and from Generalitat Valenciana (GVA) under project \emph{DeepPattern} (grant agreement PROMETEO/2019/121). We gratefully acknowledge the support of NVIDIA Corporation with the donation of a GPU used for part of this research, and Andr{\'e}s Trapiello and Ediciones Destino for granting us permission to use their book in our research. Additionally, we would like to thank all the volunteers that took part in the user study, and the scholars from \emph{Prolope} that took part in the human evaluation.

\bibliographystyle{model2-names}
\bibliography{modernization}

\end{document}